\pdfoutput=1
\documentclass[letterpaper, 10 pt, conference]{ieeeconf}
\IEEEoverridecommandlockouts
\usepackage{amsmath,amssymb}
\usepackage{graphicx}
\usepackage{booktabs}
\usepackage{url}
\usepackage{xcolor}
\usepackage[hidelinks]{hyperref}

\title{\LARGE \bf Gating Before Commitment: Anticipating Intent Divergence to Prevent Post-Interaction Decision Failures in Autonomous Driving}

\author{Cong Xu$^{1}$ and Ravi Sankar$^{1}$%
\thanks{$^{1}$Cong Xu and Ravi Sankar are with the iCONS Lab, Department of Electrical Engineering, University of South Florida, Tampa, FL 33620, USA.
{\tt\small \{cong, sankar\}@usf.edu}}%
}

\begin{document}
\maketitle
\thispagestyle{empty}
\pagestyle{empty}

\begin{abstract}
Intent misinterpretation during vehicle interactions causes recurring planning failures. We study a decision layer in which a language-guided intent module reads structured descriptors, computes a smoothed intent-geometry divergence score, and gates the planned maneuver before commitment, upstream of a corridor envelope. On a replayed off-road departure and four crash clips under a frozen, disclosed implementation, gating is the only layer that repairs the plan: on the main case it fires 72~ms after the drift onset but 161~ms before the corridor exit, keeping the trajectory in the corridor in all ten replays. The first calibration draws nine false triggers in 5.9 minutes, each from scoring uncertainty as half a conflict; a preregistered redesign treating uncertainty as abstention cuts this to 0.341 per minute. Two ablations bound the model's contribution: the full score detects fastest on four of five failures under the deployed eligibility, three of five against the unvetoed rule (000871 by one cycle; 000228 by a pre-onset fire on an uncertain stretch that five clips cannot classify as signal or coincidence; dropping the confidence term costs two detections), while on in-domain tracks at equal false positives the geometric rule more than triples its detection. The evidence supports the gating mechanism; the model's demonstrated roles are the fastest detection on these failures and an uncertainty veto on the geometric rule.
\end{abstract}

\section{Introduction}

Autonomous vehicles routinely interact with surrounding traffic, passing oncoming vehicles, negotiating merges, and responding to lane changes, yet these interaction phases remain a primary source of catastrophic decision failures~\cite{nhtsa2023,koopman2016}. Such failures are usually attributed to perception limits. Perception does degrade, including in the incident replayed here, where lane estimates are disturbed for a few frames as the oncoming vehicle clears. The question this paper asks is what the decision layer should do in that window: a planner that commits a lateral maneuver inconsistent with the interaction phase turns a transient perception disturbance into an unrecoverable state, and the signal we test is the consistency between the ego's own committed maneuver and the inferred interaction semantics~\cite{schwarting2018,hubmann2017}.

The post-interaction phase, the interval immediately after an oncoming vehicle passes or a merge completes, is especially dangerous. During this window the ego planner may attempt to correct a perceived lateral offset without recognizing that the correction itself constitutes an intent divergence from safe operation. In the real incident replayed in Section~\ref{sec:case}, the measurable drift begins 167~ms after the oncoming vehicle clears the ego plane, 
and the vehicle leaves the road corridor within 400~ms. 
If the drift carries the vehicle onto an unpaved shoulder, the change in tire-road friction can escalate the departure into a loss of control.

Existing safety architectures address this problem from the consequence side. Control barrier functions~\cite{ames2017}, responsibility-sensitive safety~\cite{rss2017}, and collision-avoidance constraints detect unsafe states and intervene reactively, after the decision has been committed. What is needed in addition is anticipation: detecting that the ego vehicle's intended maneuver is diverging from safe interaction semantics, and repairing the plan rather than overriding it.

We study a decision layer comprising (1) a language-guided multi-agent intent anticipation module that infers intent hypotheses, computes an intent-geometry divergence score, and triggers maneuver gating and plan repair; (2) a monocular perception stack and a lightweight digital twin that provide scene context; and (3) an intent-conditioned safety envelope as a reactive backstop. This paper reports two experimental rounds against one frozen implementation. Round one (v1) measures the mechanism and diagnoses a specific scoring defect. Round two (v2), preregistered before implementation, redesigns the uncertainty scoring and re-measures everything. The contributions of the paper are:
\begin{itemize}
\item A framing of a class of decision failures in interactive driving as intent-level failures, misinterpretations of ego and other-vehicle intent, rather than solely perception failures.
\item A fully disclosed implementation (base model, adapter, rule-generated training data, calibration) and a replay protocol whose instant definitions reproduce every latency number, applied to a real failure, four public crash clips, and a nominal-driving corpus.
\item A diagnosis of the uncertainty-scoring failure: scoring the model's abstentions as half a conflict produced all nine nominal false triggers, and, as the cycle logs later showed, one coincidental detection.
\item An abstention-semantics redesign with dual-threshold hysteresis, preregistered with acceptance criteria, that removes most of the false-trigger cost (1.54 to 0.341 per minute) while preserving every intent-level detection; the one criterion it misses is reported as failed and anatomized.
\item Two ablations that bound our own headline component, with a disclosed correction (a replay defect had reported identical firing cycles for every scoring variant): the corrected replay shows the full score detects fastest on four of five failures under the deployed eligibility, three of five against the unvetoed rule (000871 by one cycle; 000228 by a pre-onset fire on an uncertain stretch), with the model's uncertain-label veto suppressing the model-free rule's nominal false triggers; on in-domain tracks the geometric rule more than triples the model's detection once false-positive rates are matched.
\end{itemize}

Section~\ref{sec:related} reviews related work. Section~\ref{sec:framework} presents the framework, the implementation, and the v2 redesign. Section~\ref{sec:case} presents the replay protocol and the main-case results. Section~\ref{sec:extended} reports both rounds of extended evidence. Section~\ref{sec:limits} discusses limitations and Section~\ref{sec:conclusion} concludes. A supplementary video shows the main-case replay (Baseline versus gated) with the divergence trace.

\section{Related Work}
\label{sec:related}

\subsection{Intent Prediction and Decision Correction}
Trajectory forecasting~\cite{gupta2018,ivanovic2019}, game-theoretic planning~\cite{wang2021game} and interaction-aware predictors~\cite{gao2020vectornet,salzmann2020} predict where vehicles will be, but output point predictions rather than intent hypotheses with uncertainty, and none closes the loop from prediction to decision correction: they inform the planner without gating or repairing its output~\cite{schwarting2018,hubmann2017}. On the failure side, scenario analysis~\cite{nhtsa2023,koopman2017}, formal verification~\cite{rss2017} and minimum-risk fallbacks~\cite{iso22737} address the aftermath rather than the anticipation, and online plan repair~\cite{pek2019} reacts to observed deviations without a model of intent divergence as the cause. Barrier-function filters with predictor-corrector loops~\cite{ames2017} add short-horizon prediction to a reactive filter; the approach here places the prediction at the intent level, upstream of the geometric filter; Section~\ref{sec:case} measures one consequence: the geometric filter flags a fast excursion earlier, the intent layer repairs it.

\subsection{Language Models for Driving Reasoning}
Large language models have been applied to driving for scenario narration~\cite{sha2023}, commonsense reasoning~\cite{driess2023}, and high-level decision support~\cite{chen2024}. The mechanism studied here positions the language model as an intent anticipation module that operates upstream of the planner and issues correction signals rather than control commands. The structured, text-only interface follows the serialization result of~\cite{xu2026serialization}, which showed that a deterministic semantic serialization of a frozen perception stack gives a text-only reader a usable perception interface at small model scale; that is what makes a 0.5B-parameter fine-tuned model viable in a planning loop.

\subsection{Scene Representation}
Dense monocular reconstruction~\cite{matsuki2024} and layered language-vision maps~\cite{xu2026lvcslam1} can supply richer context than the descriptors used here, and the implementation measured in this paper contains neither: its decision loop consumes a detector, tracker, lane estimator and pinhole ground-plane model (Section~\ref{sec:impl}), with the bird's-eye corridor derived from the same lane fits; every number below characterizes that lean stack. A complementary line of work sustains the same structured state when the visual channel degrades~\cite{xu2026absence}; this paper asks what the decision layer should do once intent becomes uncertain, and v2 answers concretely: treat uncertainty as abstention, not conflict.

\section{Framework}
\label{sec:framework}

\begin{figure*}[t]
\centering
\includegraphics[width=0.92\textwidth]{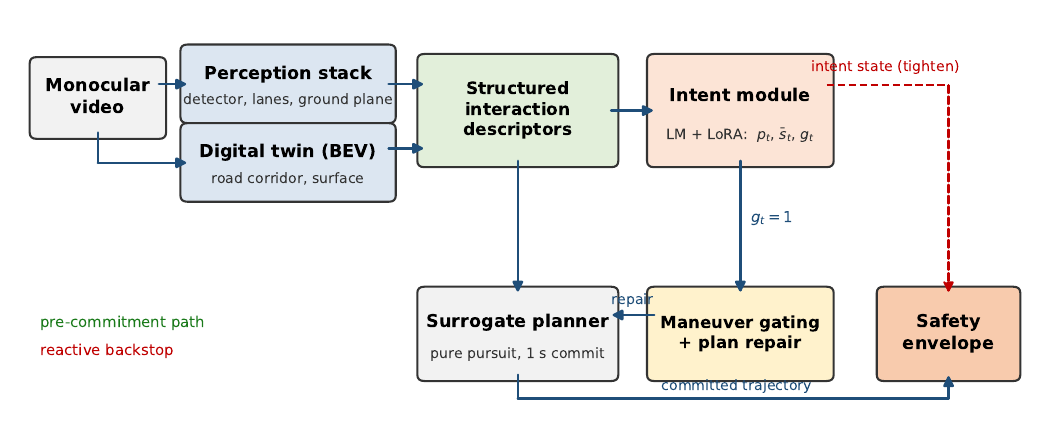}
\caption{System overview. Monocular video feeds the perception stack and a lightweight digital twin, which provide scene context for the language-guided intent anticipation module. Intent hypotheses and the divergence signal drive maneuver gating and plan repair upstream of trajectory commitment. A safety envelope, conditioned on the intent state, bounds residual risk as a backstop.}
\label{fig:overview}
\end{figure*}

\begin{figure*}[t]
\centering
\includegraphics[width=0.92\textwidth]{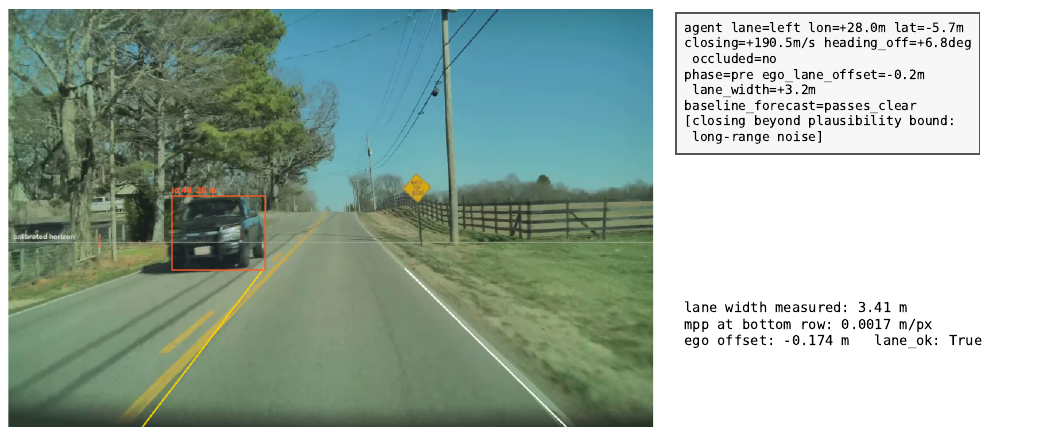}
\caption{The rebuilt perception stack on main-case frame 968: YOLOv8 detection with ByteTrack id, color-based centerline/edge-line fits, calibrated horizon, ground-plane range, and the serialized descriptor record consumed by the intent module. The record carries the per-frame lane width (3.2~m here) while the side panel reports the clip-level median used to anchor the ground plane (3.41~m); the descriptor is deliberately per-frame. The closing-speed entry exceeds the plausibility bound at this range and is marked as long-range noise (Section~\ref{sec:fma}).}
\label{fig:pipeline}
\end{figure*}

Fig.~\ref{fig:overview} illustrates the architecture; Fig.~\ref{fig:pipeline} shows the perception stack's actual outputs on a main-case frame.

\subsection{Perception and Scene Memory}
The decision layer operates on monocular RGB video. Dynamic objects are detected and tracked as bounding-box agents (YOLOv8s with ByteTrack), lane boundaries are estimated by color-based fits of the centerline and edge line, and a pinhole ground-plane model anchored on the per-clip lane-line vanishing point converts image positions to metric range and lateral offset. A lightweight digital twin renders a bird's-eye-view map whose road corridor is annotated semi-automatically: the lane width is measured from the detected boundary lines (3.41~m on the main case) 
and the corridor is centered on the lane with a fixed half-width. This scene memory supplies the structured interaction descriptors fed to the intent module.

\subsection{Language-Guided Multi-Agent Intent Anticipation}
\label{sec:intent}
For each interaction participant the intent module infers intent hypotheses with confidence estimates and an intent divergence signal. Inputs are structured descriptors: relative positions, closing speed, heading offset, lane assignment, an occlusion flag, and the interaction-phase label (pre-interaction, during, post-interaction). On the monocular clips the occlusion flag is set only for an agent that has passed out of the field of view; the nuScenes windows use the recorded visibility level. Outputs are intent hypotheses with uncertainty and correction/gating signals; the model never outputs control commands and no raw pixels enter it. Each cycle carries a risk-intent state for the primary interacting agent: Red (intent conflict), Yellow (advisory), and Green (aligned). Red issues the correction signal.

\subsubsection{Intent distribution and uncertainty}
Let the interaction at planning cycle $t$ be summarized by a descriptor vector $x_t$. The module maps $x_t$ to a hypothesis over $\mathcal{I} = \{\text{yield}, \text{block}, \text{merge}, \text{pass}, \text{uncertain}\}$ as a categorical distribution
\begin{equation}
p_t(i) \triangleq \Pr(I_t = i \mid x_t), \qquad
p_t(i) = \frac{\exp\!\big(z_t(i)/\tau\big)}{\sum_{j} \exp\!\big(z_t(j)/\tau\big)},
\end{equation}
where $z_t$ are the model's unnormalized label scores and $\tau > 0$ calibrates confidence. The MAP label is $\hat{\imath}_t = \arg\max_{i} p_t(i)$ and the confidence is $c_t \triangleq p_t(\hat{\imath}_t)$; $c_t$ is the only distributional quantity the gate consumes.

\subsubsection{Intent-geometry divergence}
Let $\pi^{\mathrm{pred}}_t$ denote the motion mode implied by the geometric predictor and the planner's currently committed maneuver set (opens\_gap, closes\_gap, crosses\_lane, passes\_clear, ambiguous; a committed lateral excursion of the ego maps to crosses\_lane), and let $\pi(\hat{\imath}_t)$ denote the semantic mode of the inferred label. Their consistency is a fixed lookup table $d_t \in [0,1]$, part of the frozen configuration. 
The divergence score and its smoothing are
\begin{equation}
s_t \triangleq (1 - c_t) + \lambda\, d_t, \qquad
\bar{s}_t \triangleq \alpha\, \bar{s}_{t-1} + (1-\alpha)\, s_t,
\label{eq:div}
\end{equation}
and the conflict flag is
\begin{equation}
g_t \triangleq \mathbb{1}\{\bar{s}_t \ge \theta\}.
\label{eq:gate}
\end{equation}

\subsubsection{Uncertainty as abstention (v2)}
\label{sec:abstention}
The first calibration (v1) scored $d(\pi^{\mathrm{pred}}, \text{uncertain}) = 0.5$ for every non-ambiguous forecast: not knowing was treated as half a conflict. Section~\ref{sec:false} shows this single entry produced every nominal false trigger, and Section~\ref{sec:clips} shows it also manufactured one apparent detection. The redesigned scoring, preregistered before implementation with acceptance criteria (Section~\ref{sec:prereg}), treats \emph{uncertain} as an abstention rather than a semantic hypothesis:
\begin{itemize}
\item $d(\pi^{\mathrm{pred}}, \text{uncertain}) = 0$ for every non-ambiguous $\pi^{\mathrm{pred}}$; the ambiguous entry was already $0$ in v1, so under v2 abstention never contributes disagreement;
\item a cycle whose MAP label is \emph{uncertain} raises at most a Yellow advisory; Red, and therefore gating and repair, is reachable only on a cycle whose own MAP label is a semantic hypothesis (yield, block, merge, pass). Abstaining cycles still contribute their $(1-c_t)$ to $\bar{s}_t$, so the memory of an uncertain stretch can carry into the first semantic cycle that follows;
\item dual-threshold hysteresis: the flag is raised when $\bar{s}_t \ge \theta_{\mathrm{on}}$ on an eligible cycle and clears only when $\bar{s}_t < \theta_{\mathrm{off}} = \theta_{\mathrm{on}} - 0.10$, the offset fixed in the preregistration.
\end{itemize}
Everything else in Eq.~\eqref{eq:div}--\eqref{eq:gate}, including the $(1-c_t)$ term, is unchanged. v2a denotes abstention alone; v2b (primary) adds hysteresis.

\subsection{Intent-Aware Decision Correction and Plan Gating}
When $g_t = 1$, the correction module intervenes upstream of commitment: maneuver gating restricts the admissible maneuver set (suppressing lateral corrections in the post-interaction window), and plan repair returns the committed trajectory toward the lane center at a bounded lateral rate. Both act before execution; their timing is measured, not assumed, in Section~\ref{sec:latency}.

\subsection{Intent-Conditioned Safety Envelope}
A safety envelope enforces road-surface containment: a committed next-horizon waypoint outside the corridor rejects the trajectory, and the corridor tightens under Yellow or Red. In the surrogate the rejection is instrumented as a signal only (no minimum-risk maneuver executes), so the envelope arm measures when a corridor check would notice the excursion, not what a controller would do about it; its 10/10 exits in Table~\ref{tab:outcome} follow from that instrumentation, not from corridor filters in general.

\subsection{Implementation, Training, and Calibration}
\label{sec:impl}
The original implementation of this system was lost; everything below describes a new implementation built in August 2026, and all numbers in this paper are measurements from it. The v2 redesign was preregistered before implementation, and $\theta_{\mathrm{on}}$ was the only recalibrated parameter. 

\textit{Model.} The intent module is Qwen2.5-0.5B (496M parameters including the adapter), 
adapted by LoRA (rank 16, alpha 32, q/k/v/o projections) 
to map a serialized descriptor string to an intent label; label scores are read from the five label-token logits, with no free-form generation. Inference runs in bfloat16 on a single RTX~5090 laptop GPU. 

\textit{Descriptor serialization.} Each $x_t$ is a fixed-schema text record (per agent: relative longitudinal and lateral position, closing speed, heading offset, lane assignment, occlusion flag, phase label, then $\pi^{\mathrm{pred}}_t$), following the scoped serialization of~\cite{xu2026serialization}; an actual record appears in Fig.~\ref{fig:pipeline}. 

\textit{Training data.} Fine-tuning uses 55{,}433 interaction windows (majority class capped) from nuScenes v1.0-trainval metadata~\cite{caesar2020nuscenes}, 
labeled by a deterministic rule on each agent's future annotated kinematics over 2.0~s: an agent is merge if it enters the ego lane while closing laterally, pass if its relative longitudinal position changes sign while it stays out of the ego lane, yield if it decelerates by at least 1.5~m/s while the gap does not shrink, block if it holds the ego corridor ahead at constant speed with the gap not growing, and uncertain otherwise. No human annotation is involved. Distribution: uncertain 30{,}296, pass 15{,}148, block 4{,}572, merge 4{,}452, yield 965. 
Windows are split by scene (80/20). 

\textit{Calibration.} $\tau = 0.885$ is fit by NLL minimization on the validation split; validation accuracy is 0.889 with ECE 0.008 on 4{,}000 held-out-scene windows; 
per-label accuracy is uneven (pass 0.90, block 0.87, uncertain 0.92, merge 0.23, yield 0.05 on small counts). 
v1 gating parameters: $\lambda = 1.0$, $\alpha = 0.7$, $\theta = 0.55$, chosen on validation interaction tracks (394 conflict vs 1{,}863 others) at the maximum of detection rate minus false-positive rate. 
For v2, the same tracks, criterion, and stored logits were used in a single pass to choose $\theta_{\mathrm{on}} = 0.35$ (detection 0.632, false positive 0.345); $\tau$, $\lambda$, $\alpha$ carry over unchanged. 
Each configuration was frozen before its experiments and untouched afterwards.

\textit{Cycle budget.} The planning cycle is 100~ms. 
The intent forward pass takes 32.7~ms median (34.2 at p90) over 200 timed calls; 
the intent-plus-gating-plus-repair portion of a cycle averages about 38~ms wall time. 
Frames are preprocessed offline; perception time is not part of any reported latency.

\section{Replay Study of a Real-World Failure}
\label{sec:case}

\subsection{Scenario}
We study a real-world monocular video of an oncoming-vehicle interaction on a two-lane rural road~\cite{syngates2025}, re-acquired at 2896$\times$1876, 30~fps, 45.75~s. 
An oncoming pickup passes and, in the original incident, the production planner generates a post-interaction lateral correction that crosses the centerline, leaves the road, and ends in a rollover. We do not have access to the production planner; the failure is replayed as described next.

\subsection{Replay Protocol, Anchors, and Latency Measurement}
\label{sec:latency}
The perception stack produces per-frame descriptors; a surrogate ego planner (pure pursuit on the perceived lane center with a lateral-offset correction term) is driven by the same descriptors, and its committed next-horizon trajectory follows the perceived lateral state. Run without correction it reproduces the observed off-road excursion; this is the Baseline.

All instants are defined by deterministic rules: 
\begin{itemize}
\item $t_0$: first frame after the oncoming vehicle's rear edge clears the ego plane; frame 978 ($t = 32.60$~s).
\item \textbf{Anchor 1, drift onset}: first frame at which the lane-normalized ego offset exceeds 1.0~m and keeps growing for five valid frames; frame 983, 167~ms after $t_0$.
\item \textbf{Anchor 2, corridor exit}: first cycle at which the Baseline's executed lateral state leaves the corridor; cycle 330 ($t = 33.0$~s), 400~ms after $t_0$. 
\item Commitment: the planner hands its next 1.0~s horizon to the tracker at the end of every 100~ms cycle; $t_{\mathrm{gate}}$ / $t_{\mathrm{env}}$: first cycle at or after $t_0$ with $g_t = 1$ / with an envelope rejection.
\end{itemize}
Every lead or offset names its anchor. Latency after $t_0$ is cycle offset plus that cycle's measured compute time, wall clock, over 10 replays per configuration. The replays are deterministic in their decisions: every replay of a configuration fires in the same cycle, and the reported standard deviations reflect only the wall-clock jitter of that cycle's computation.

\subsection{Configurations and Results}
We compare Baseline, Intent Only, Envelope Only, and Full, each under v1 and under the preregistered v2 configurations.

\begin{table}[t]
\caption{Main-case outcome and detection latency, 10 replays per row; $\pm$ is one standard deviation. Gate anchors: drift onset (A1) and corridor exit (A2).}
\label{tab:outcome}
\centering
\small
\renewcommand{\arraystretch}{0.95}
\setlength{\tabcolsep}{2.6pt}
\begin{tabular}{llccc}
\toprule
Cfg. & Arm & Exit & Repaired & Latency after $t_0$ \\
\midrule
v1  & Baseline      & 10/10 & no  & n/a \\ 
v1  & Envelope Only & 10/10$^{\dagger}$ & no  & $100.0 \pm 0.0$~ms \\ 
v1  & Intent Only   & 0/10  & yes & $238.1 \pm 1.1$~ms \\ 
v1  & Full          & 0/10  & yes & $239.6 \pm 2.3$~ms \\ 
v2a & Intent / Full & 0/10 & yes & $233.3 \pm 1.0$~ms \\ 
v2b & Intent / Full & 0/10 & yes & $234.4 \pm 3.7$~ms \\ 
\bottomrule
\end{tabular}

\vspace{2pt}
\parbox{\columnwidth}{\footnotesize $^{\dagger}$The envelope rejects the committed trajectory in 10/10 replays (interception) but acts after commitment and does not modify the plan. Baseline and Envelope arms are gate-independent and identical across configurations. 
}
\end{table}

Table~\ref{tab:outcome} reports the outcome. The Baseline exits the corridor in all replays. Intent gating repairs the plan in all replays under v1, v2a, and v2b alike: relative to the two anchors, the gate fires $72.2 \pm 2.0$~ms after the drift onset and $161.1 \pm 2.0$~ms before the corridor exit under v1, and $67.7 \pm 3.7$~ms after / $165.6 \pm 3.7$~ms before under v2b, always before the corresponding commitment. 
The envelope flags the excursion at $100.0 \pm 0.0$~ms after $t_0$ (its horizon check spans the full committed 1~s, so on a fast excursion it is effectively a one-cycle detector), 
but only the gated planner produces a corrected, in-corridor commitment.

\begin{figure}[t]
\centering
\includegraphics[width=\columnwidth]{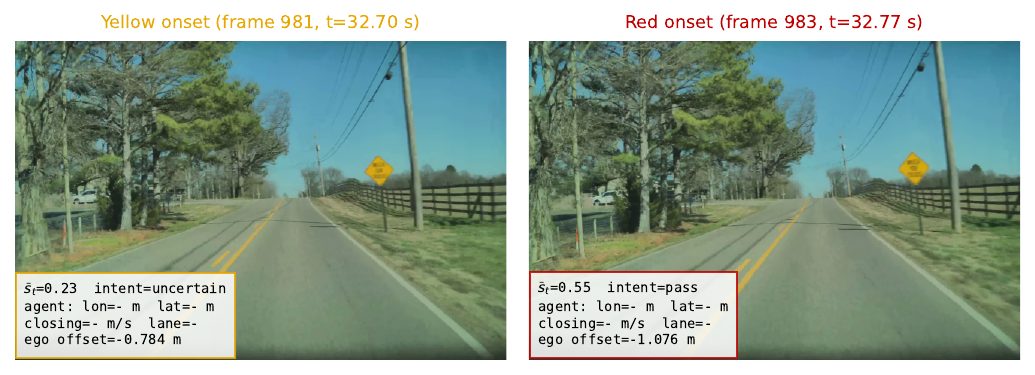}
\caption{Main case, v2b Full replay: the input frames of the Yellow-onset and Red-onset cycles (frames 981 and 983), not the instants at which those states take effect; the cycle that consumes frame 983 starts at $t_0 + 200$~ms and its gate decision lands 34~ms into that cycle, which is the $t_0 + 234$~ms of Table~\ref{tab:outcome}. The panel shows dashes when no agent is in view, which is the situation throughout the post-interaction window. Closing speed beyond the plausibility bound is displayed as long-range noise (Section~\ref{sec:fma}). All quantities from the recorded cycle log. 
}
\label{fig:case}
\end{figure}

\subsection{Failure Mode Analysis}
\label{sec:fma}
The anatomy is identical in all replays. 
At $t_0$ the passed vehicle leaves the field of view; one cycle later the planner, following the corrupted post-pass lane perception, commits a horizon that crosses the corridor bound, which the envelope flags immediately. One cycle after that the committed lateral rate becomes geometrically abnormal (crosses\_lane), the intent module reads a low-confidence \emph{pass} ($c_t = 0.51$) against it, the table scores $d_t = 0.8$, and $\bar{s}_t$ crosses the threshold: Red, gate, repair. 
Because the trigger routes through a semantic hypothesis, not through \emph{uncertain}, the v2 scoring leaves it intact, as Table~\ref{tab:outcome} confirms.

One perception observation belongs here: beyond roughly 50~m ground-plane ranging is pixel-noise dominated and serialized closing speeds become implausible (a v1 cycle carried 299.7~m/s); this sits upstream of some \emph{uncertain} outputs, tying perception noise to the abstention pathway of Section~\ref{sec:abstention}, and the figures mark such values rather than displaying them as physics. The frozen serializer does not clip them; clipping belongs to the next freeze. 

\begin{figure}[t]
\centering
\includegraphics[width=\columnwidth]{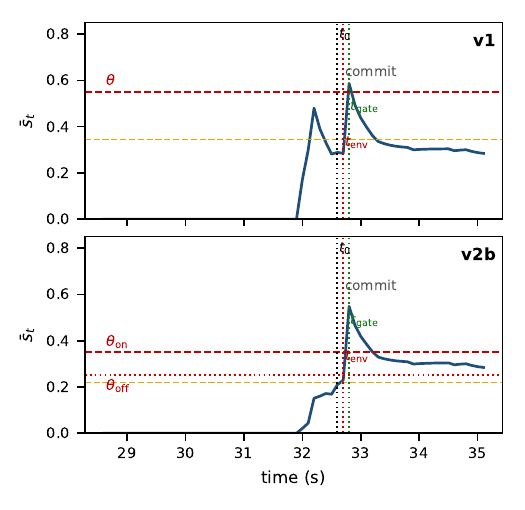}
\caption{Smoothed divergence $\bar{s}_t$ on the main case for v1 (top) and v2b (bottom), with $t_0$, the first commitment, $t_{\mathrm{gate}}$, $t_{\mathrm{env}}$, and for v2b the hysteresis pair $\theta_{\mathrm{on}}/\theta_{\mathrm{off}}$. The v2b trace stays above $\theta_{\mathrm{off}}$ to the end of the window, so the flag remains raised and lateral corrections stay gated for the rest of the replay; release requires $\bar{s}_t$ to fall below $\theta_{\mathrm{off}}$, which needs a run of confident, geometry-consistent cycles. From the recorded cycle logs. 
}
\label{fig:timeline}
\end{figure}

\section{Two Rounds of Extended Evidence}
\label{sec:extended}

\subsection{Preregistration and Acceptance}
\label{sec:prereg}
The v2 redesign was preregistered before implementation, with the acceptance criteria fixed in advance and $\theta_{\mathrm{on}}$ declared the only recalibrated parameter. 
Table~\ref{tab:prereg} reports the verdicts. A2 is reported as \textbf{failed as written} and is not redefined retroactively: its wording asks for a flag before the drift onset in all five cases, and two cases miss that instant. Clip 000228 misses it by 1.1~s (Section~\ref{sec:clips}). The main case misses it by $67.7 \pm 3.7$~ms: its gate falls after the drift onset, though it precedes the corridor exit by $165.6 \pm 3.7$~ms (Section~\ref{sec:latency}), which is the anchor the criterion should have named. v1 misses the same criterion on the main case by $72.2 \pm 2.0$~ms. 
A mixed-anchor criterion would have passed 4/5; we did not rewrite it. The dated adjudication addendum keeps v2b primary and requires the failed item disclosed and anatomized. 

\begin{table}[t]
\caption{Preregistered acceptance criteria for v2b and measured verdicts.}
\label{tab:prereg}
\centering
\small
\setlength{\tabcolsep}{3.0pt}
\begin{tabular}{llc}
\toprule
 & Criterion (preregistered) & Verdict \\
\midrule
A1 & false triggers $\le 0.5$/min & pass: 0.341/min \\ 
A2 & flag before drift onset, all 5 cases & \textbf{fail: 3/5} \\ 
A3 & E1 outcomes unchanged & pass \\ 
A4 & merge control (descriptive) & Red 6/10 $\to$ 4/10 \\ 
\bottomrule
\end{tabular}
\end{table}

\subsection{Additional Interaction Clips}
\label{sec:clips}
From the 801 ego-involved crash clips of the Car Crash Dataset~\cite{bao2020ccd} we screened for clips with (F1) a visible pass/merge/overtake/oncoming interaction before the accident, (F2) failure onset within 3~s of the interaction end, and (F3) an ego-motion failure rather than third-party intrusion. Four clips passed; at screening freeze each was typed: 000754 and 000871 (with the main case) are \emph{intent-level} failures (the ego commits a wrong maneuver against the interaction semantics), while 000676 and 000228 are post-interaction \emph{loss-of-control boundary checks}. Every screened clip's accept or reject decision, with the failing flag, is recorded in the artifact bundle.

\begin{table}[t]
\caption{Divergence-flag behavior on the four CCD clips under v1 and v2b. Offset is from $t_0$ (gate-cycle compute included); lead is to the drift-onset anchor; offset plus lead is therefore constant per clip up to compute time. Negative lead = flag after onset.}
\label{tab:clips}
\centering
\small
\renewcommand{\arraystretch}{0.95}
\setlength{\tabcolsep}{2.8pt}
\begin{tabular}{llcrrrr}
\toprule
 &  &  & \multicolumn{2}{c}{v1} & \multicolumn{2}{c}{v2b} \\
Clip & Type & Level & Offset & Lead & Offset & Lead \\
\midrule
000754 & oncoming & intent & 436 & 700 & 133 & 1000 \\ 
000871 & overtake & intent & 137 & 1400 & 138 & 1400 \\ 
000676 & pass & boundary & 838 & 800 & 834 & 800 \\ 
000228 & pass & boundary & 1034 & 100 & 2234 & $-1100$ \\ 
\bottomrule
\end{tabular}
\end{table}

Under v1 the uncorrected planner exits the corridor in 4/4 and the flag precedes the drift onset in 4/4 (median lead 750~ms). 
Under v2b, every \emph{intent-level} failure is detected with equal or better lead (000754: 700 to 1000~ms; 000871: 1400~ms; main case unchanged), and boundary check 000676 is retained (800~ms). Boundary check 000228 is the preregistered criterion v2 fails: its flag comes 1.1~s after the onset. The cycle logs show why: the v1 trigger on this clip was reached with 10 of the 12 preceding cycles carrying MAP \emph{uncertain}, 9 of them scored $d = 0.5$, the same anatomy as the nine nominal false triggers of Section~\ref{sec:false}. 
The v1 detection on 000228 was the diagnosed defect firing coincidentally 100~ms before the onset. In one sentence: the defect inflated both the false positives and the apparent detections, and the fix converges both.

On these 5~s pre-crash cuts the ego is already at the corridor boundary when the gate fires, so repair mitigates but cannot restore containment (0/4 both rounds); detection lead and recoverability remain separate quantities. 

\subsection{False-Trigger Rate on Nominal Driving}
\label{sec:false}
Zero-interaction scenes are essentially absent in urban nuScenes: one clear-weather validation scene (0.06~min) contains no event of any kind. The nominal set is therefore a disclosed proxy: the 20 largest clear-weather validation scenes with no conflict event (no block/merge/yield window); routine passes remain. 
Cycles are keyframe-synchronous at 2~Hz, against 10~Hz on the clips and the highway probe. With $\alpha = 0.7$ the smoother's time constant is therefore five times longer in wall-clock terms on this set, which makes a trigger harder to reach per unit time; the per-minute rates across sampling rates are indicative rather than strictly comparable.

\begin{table}[t]
\caption{False triggers on the identical 703-cycle (5.86~min) nominal set, and on the out-of-domain highway probe$^{\ddagger}$ (8 comma2k19 segments, 4{,}792 cycles, 8.0~min; reported separately, not merged).}
\label{tab:false}
\centering
\small
\begin{tabular}{lcccc}
\toprule
Config & Triggers & Per minute & Yellow & Red \\
\midrule
v1  & 9 & 1.537 & 30.9\% & 5.8\% \\ 
v2a & 2 & 0.341 & 5.3\% & 0.3\% \\ 
v2b & 2 & 0.341 & 4.1\% & 1.4\% \\ 
\midrule
v1 (hwy)$^{\ddagger}$ & 203 & 25.4 & 16.2\% & 45.5\% \\ 
v2b (hwy)$^{\ddagger}$ & 60 & 7.5 & 18.7\% & 19.9\% \\ 
\bottomrule
\end{tabular}
\end{table}

Table~\ref{tab:false} reports both rounds. All nine v1 triggers share one anatomy: a routine pass is labeled \emph{uncertain} with moderate confidence, scored as half a conflict, and the EWMA integrates past $\theta = 0.55$. 
As a diagnostic bridge (a post-hoc replay), zeroing that single table entry on the stored logits removes all nine. 
The fresh v2 runs land at 2 triggers rather than 0 because $\theta_{\mathrm{on}}$ was recalibrated down to 0.35: both v2b triggers route through semantic \emph{pass} cycles rather than through \emph{uncertain}, and both need the $(1-c_t)$ term to clear $\theta_{\mathrm{on}}$: one at $c_t = 0.53$ against an unremarkable forecast, and one on a single crosses\_lane cycle at $c_t = 0.76$, where the disagreement term alone reaches only $0.24$ after smoothing. 
The remaining rate, 0.341 per minute, is within the preregistered bound; the abstention alone (v2a) accounts for the reduction, and hysteresis mainly stabilizes the raised flag (Red fraction 0.3\% to 1.4\% as the flag persists between thresholds). 

As an out-of-domain probe, eight comma2k19 highway segments~\cite{schafer2018comma} (first segment per route, fixed luma rule, no content-based selection) ran through the identical frozen pipeline: 4{,}792 cycles, 8.0~minutes. Both configurations far exceed the usability bound (v1 25.4, v2b 7.5 per minute): the lane estimator keys on a yellow centerline this corpus lacks (fits on 4.5\% of frames), the ground plane falls back to a default horizon, and 44 of 60 v2b triggers pair a \emph{block} hypothesis (median confidence 0.84) with an ambiguous forecast in dense car-following. These are semantic-route triggers from out-of-domain perception, not the repaired scoring; they quantify the calibration-domain limitation and are reported separately. 

\subsection{What the Language Model Contributes}
\label{sec:ablation}
The main-case trigger routes through a \emph{pass} hypothesis read against a crosses\_lane forecast, and the forecast mode is set by the ego's own committed lateral rate. How much of the detection is the language model doing? We answer with a diagnostic replay that varies only the scoring term and holds every other configuration element fixed: \emph{full} is the paper's score; \emph{d-only} drops the $(1-c_t)$ term; \emph{geometry-only} ignores the model entirely and scores $1$ whenever $\pi^{\mathrm{pred}}_t = \text{crosses\_lane}$. The indicator passes through the same EWMA and the frozen $\theta_{\mathrm{on}} = 0.35$, so a single crosses\_lane cycle contributes $1 - \alpha = 0.3$ to $\bar{s}_t$.

A defect disclosure first: an earlier run of this ablation reported all three variants firing in identical cycles, because the replay driver silently ignored the variant switch and ran the full score three times; the defect was exposed by a later fusion experiment and the driver corrected. Tables~\ref{tab:outcome}, \ref{tab:clips} and \ref{tab:false} were unaffected: the first two run only the full score, and the nominal-set path had always produced distinct per-variant counts. The one-line fix and an independent recomputation of every variant cycle from the logged inputs are in the artifact bundle; the recomputation reproduces the corrected driver cycle for cycle. 
The corrected replay reverses that reading: the full score detects fastest on four of five failures under the deployed eligibility (three of five against the unvetoed rule, below) and most completely. Under the deployed eligibility, geometry-only fires on all five but one cycle later on the main case, 300~ms later on 000754, after the onset on both boundary checks (000676 at $-1.0$~s where full leads by $0.8$~s; 000228 at $-1.5$~s against full's $-1.1$~s), and one cycle before full only on 000871; $d$-only never fires on either boundary check, so dropping the $(1-c_t)$ term costs two of the five detections. Under pure, model-free eligibility the rule moves only on the boundary checks, to cycle 15 (000676) and 12 (000228), both \emph{before} the onset: the abstention veto pushes both past it, suppressing pre-onset geometric triggers that ride \emph{uncertain} stretches, v1's coincidental-detection anatomy; signal or coincidence, five clips cannot say. 
The abstention veto acts alongside: the gate applies it (an \emph{uncertain} cycle cannot raise Red) to every variant, and the genuinely model-free rule, freed of it, draws four nominal false triggers (0.68 per minute) against the full score's two and $d$-only's none; the veto removes all four and none of the deployed gate's five detections, each of which fires on a semantic-label cycle, while delaying the unvetoed rule past the onset on both boundary checks. Because \emph{uncertain} is a 0.92-accuracy majority class, the veto may still be blanket rather than selective on this evidence. 
All five cases contain a geometrically abnormal ego maneuver by the time the flag is due; the regime where the conflict is legible in the other agent first is tested in Section~\ref{sec:antic}.

\subsection{Anticipation When the Cue Is in the Other Agent}
\label{sec:antic}
To test the regime the replays cannot reach, we evaluate the same scoring variants on nuScenes interaction tracks, where there is no ego planner and therefore no committed-maneuver cue. Tracks are the (scene, instance) sequences of the validation split; a track is positive if any of its windows carries the rule label \emph{block} or \emph{merge} (394 tracks), negative otherwise (1{,}863). 
Because $\theta$ was calibrated for the full score, each variant is swept over the same grid and reported at its own best operating point; \emph{anticipated} counts tracks whose flag rises strictly before the first conflict-labelled window.

\begin{table}[t]
\caption{Conflict anticipation on 394 positive and 1{,}863 negative nuScenes tracks, each variant at its own best point; daggered rows: both channels at matched false positives (0.093). Geometry-only rows are model-free (no abstention eligibility); anticipated = flags raised before the first conflict-labelled window.}
\label{tab:antic}
\centering
\footnotesize
\renewcommand{\arraystretch}{0.95}
\setlength{\tabcolsep}{3.0pt}
\begin{tabular}{lccccc}
\toprule
Score & $\theta$ & Det. & FP & Anticip. & Lead \\
\midrule
full & 0.35 & 0.632 & 0.345 & 67 & 3.0~s \\ 
$d$ only & 0.18 & 0.535 & 0.257 & 55 & 3.0~s \\ 
$(1-c)$ only & 0.20 & 0.726 & 0.514 & 71 & 3.5~s \\ 
geometry only & 0.10 & 0.896 & 0.213 & 45 & 2.0~s \\ 
\midrule
full$^\dagger$ & 0.78 & 0.234 & 0.093 & 21 & 1.5~s \\ 
geometry only$^\dagger$ & 0.45 & 0.800 & 0.093 & 19 & 2.5~s \\ 
\bottomrule
\end{tabular}
\end{table}

Read at each variant's own best point, Table~\ref{tab:antic} looks like a trade: the geometric rule discriminates far better (Youden $J$ of 0.683 against 0.287) while the model-driven scores anticipate more conflicts, 67 and 71 against 45, with slightly longer leads. That reading does not survive matching: anticipating is easy for a flag that fires readily, and the model-driven rows sit at up to 2.4 times the geometric rule's false positives. Held at the same 0.093 false positives, the full score detects 0.234 of the conflicts and anticipates 21 with a 1.5~s median lead, against the geometric rule's 0.800, 19 and 2.5~s. 
Splitting the positives does not rescue the model either: on the 55 tracks whose conflict is block only, and therefore longitudinal rather than lateral, the matched geometric rule still detects more (0.527 against 0.364; detection is track-level, so lateral motion elsewhere on the track counts) and neither variant anticipates a single one. 
On this evidence the language model does not buy detection or discrimination once precision is held fixed, and its anticipation edge is two tracks against a 1.0~s shorter lead. Three caveats bound that statement rather than soften it: the positive class is defined by the same rule that produced the training labels and whose lateral-velocity ingredient the geometric variant keys on, so both see part of the label's construction; the tracks are those $\theta_{\mathrm{on}}$ was calibrated on (the full row at $\theta = 0.35$ reproduces Section~\ref{sec:impl}'s calibration numbers; no held-out split is claimed); and a single rule-labelled corpus cannot speak for interactions whose semantics are not reducible to kinematics.

A preregistered control (A4) scored 10 merge scenes at frozen thresholds: Yellow rose in the merge windows in 6/10 scenes in both rounds; Red in those windows fell from 6/10 (v1) to 4/10 (v2b), so Red selectivity improves under v2 but remains partial. 

\section{Limitations}
\label{sec:limits}
The implementation was rebuilt in August 2026 after the original artifacts were lost; latency figures characterize this surrogate stack, not a production planner. v2 was evaluated on the same data as v1 by design; no new domains are claimed. The gate was calibrated on nuScenes validation tracks and the false-trigger measurements are in-domain; the main case (rural) and the CCD clips lie outside the calibration domain, no transfer of the calibration is claimed for them, and the highway probe of Section~\ref{sec:false} shows it does not transfer to out-of-domain perception. CCD clips are 5~s pre-crash cuts, too short to assess recoverability. The road corridor is semi-automatic and the monocular ground plane rests on an assumed camera height and field of view, with far-range closing speeds dominated by noise (Section~\ref{sec:fma}). The intent module's minority classes (yield, merge) are weak; labels come from a deterministic rule. The v2 recalibration touched only $\theta_{\mathrm{on}}$; the $(1-c_t)$ term is the remaining confidence-as-conflict pathway: v2 removed the route through the \emph{uncertain} label, but a merely unconfident semantic hypothesis can still drive the score, which buys two of the five detections at the cost of both surviving nominal triggers (Section~\ref{sec:ablation}). Descriptors are monocular; the cooperative (V2V) extension~\cite{xu2022opv2v} is not evaluated here.

\section{Conclusion}
\label{sec:conclusion}
We measured an intent-gating mechanism end to end under frozen, fully disclosed configurations, twice. Round one supported it as a prevention layer: on five real ego failures the flag preceded the corridor exit on the main case and the drift onset on the four clips, and gating was the only layer that repaired the plan. Round one also priced it: detection lagged a full-horizon corridor check, and nominal driving drew 1.54 false triggers per minute, all from scoring model uncertainty as half a conflict. Round two, preregistered before implementation, redesigned that single scoring choice into abstention semantics with hysteresis: false triggers fell to 0.341 per minute and every intent-level detection was preserved, while the one preregistered criterion v2 misses is precisely where the cycle logs show a v1 detection manufactured by the scoring defect: a defect that inflates false positives can inflate detections too. The same discipline then corrected our own bound: a replay defect had made the five-case ablation report identical firing cycles for every scoring variant, and the corrected replay reverses that reading. On the five real failures the full score detects fastest on four of five under the deployed eligibility, three of five against the unvetoed rule (000871 by one cycle; 000228 by a pre-onset fire on an uncertain stretch; dropping the confidence term costs two detections), and the model's uncertain veto keeps the nominal set quiet (four model-free false triggers against two). Where the cue is in the other agent instead, the model appears to buy anticipation, 67 conflicts against 45, until the comparison is held at equal false-positive rate, at which point the geometric rule roughly matches its anticipation and more than triples its detection. That is a narrow result on one rule-labelled corpus and five replayed failures; the work that would change it: interactions whose semantics kinematics cannot recover, and a score that combines the two cues rather than letting one stand in for the other.

\end{document}